\documentclass[letterpaper, 10 pt, conference]{ieeeconf}  

\IEEEoverridecommandlockouts                              

\usepackage{cite}
\usepackage{amsmath,amssymb,amsfonts}
\usepackage[linesnumbered,lined,ruled,commentsnumbered]{algorithm2e}
\usepackage{graphicx}
\usepackage{textcomp}
\usepackage{subfig}
\usepackage{hyperref}

\title{\LARGE \bf
Robust Odometry using Sensor Consensus Analysis
}

\author{Andrew W. Palmer and Navid Nourani-Vatani
\thanks{This work was supported by Siemens AG, Germany. The authors are with Siemens AG, Berlin, Germany. Email: {\tt \{andrew.palmer,navid.nourani-vatani\}@siemens.com}}}%

\begin{document}

\begin{table*}
	\copyright 2018 IEEE. Personal use of this material is permitted. Permission from IEEE must be obtained for all other uses, in any current or future media, including reprinting/republishing this material for advertising or promotional purposes, creating new collective works, for resale or redistribution to servers or lists, or reuse of any copyrighted component of this work in other works.
	
	The published version of this article can be found at https://doi.org/10.1109/IROS.2018.8594473
\end{table*}

\pagebreak

\maketitle
\thispagestyle{empty}
\pagestyle{empty}

\begin{abstract}
Odometry forms an important component of many manned and autonomous systems. 
In the rail industry in particular, having precise and robust odometry is crucial for the correct operation of the Automatic Train Protection systems that ensure the safety of high-speed trains in operation around the world. 
Two problems commonly encountered in such odometry systems are miscalibration of the wheel encoders and slippage of the wheels under acceleration and braking, resulting in incorrect velocity estimates. 
This paper introduces an odometry system that addresses these problems. 
It comprises of an Extended Kalman Filter that tracks the calibration of the wheel encoders as state variables, and a measurement pre-processing stage called Sensor Consensus Analysis (SCA) that scales the uncertainty of a measurement based on how consistent it is with the measurements from the other sensors. 
SCA uses the statistical z-test to determine when an individual measurement is inconsistent with the other measurements, and scales the uncertainty until the z-test passes. 
This system is demonstrated on data from German Intercity-Express high-speed trains and it is shown to successfully deal with errors due to miscalibration and wheel slip. 

\end{abstract}

\section{Introduction}

Precise odometry is critical to the safe operation of many manned and autonomous vehicles including autonomous underground mining vehicles \cite{Scheding1999}, planetary rovers \cite{Maimone2007}, and railways \cite{Allotta2001}. 
On-board odometry systems in the rail industry play a crucial role in the Automatic Train Protection (ATP) systems that prevent accidents due to speeding and collisions. 
The necessity for robust and accurate odometry has increased as trains travel ever faster and move towards semi- and fully-autonomous operation. 
This paper investigates robust odometry approaches that can deal with the numerous error sources that commonly present themselves on rail vehicles. 

A typical high-speed train, such as the Intercity-Express (ICE) that is operated throughout much of Europe, regularly travels at speeds of over 300km/hr, and uses a pair of wheel encoders and a pair of Doppler radars to estimate its velocity. 
In real-world operations, this sensor suite suffers from multiple error sources. 
The wheel encoders are typically mounted on driven and braked wheels, which frequently slip for long periods (10's of seconds) during acceleration and braking. 
They also rely on accurate measurement of the wheel diameter during maintenance to produce correct velocity measurements. 
In practice, the wheel diameter changes over time due to temperature fluctuations and wear, and may even be incorrectly measured in the first place. 
The Doppler radars, while unaffected by wheel slip, are known to operate poorly when the track is covered by snow or ice \cite{Lauer2015}. 
They are also inherently affected by the ground suface.

\begin{figure}[htbp]
	\centering
	\subfloat[Wheels slipping during acceleration]{
		\includegraphics[width=\linewidth]{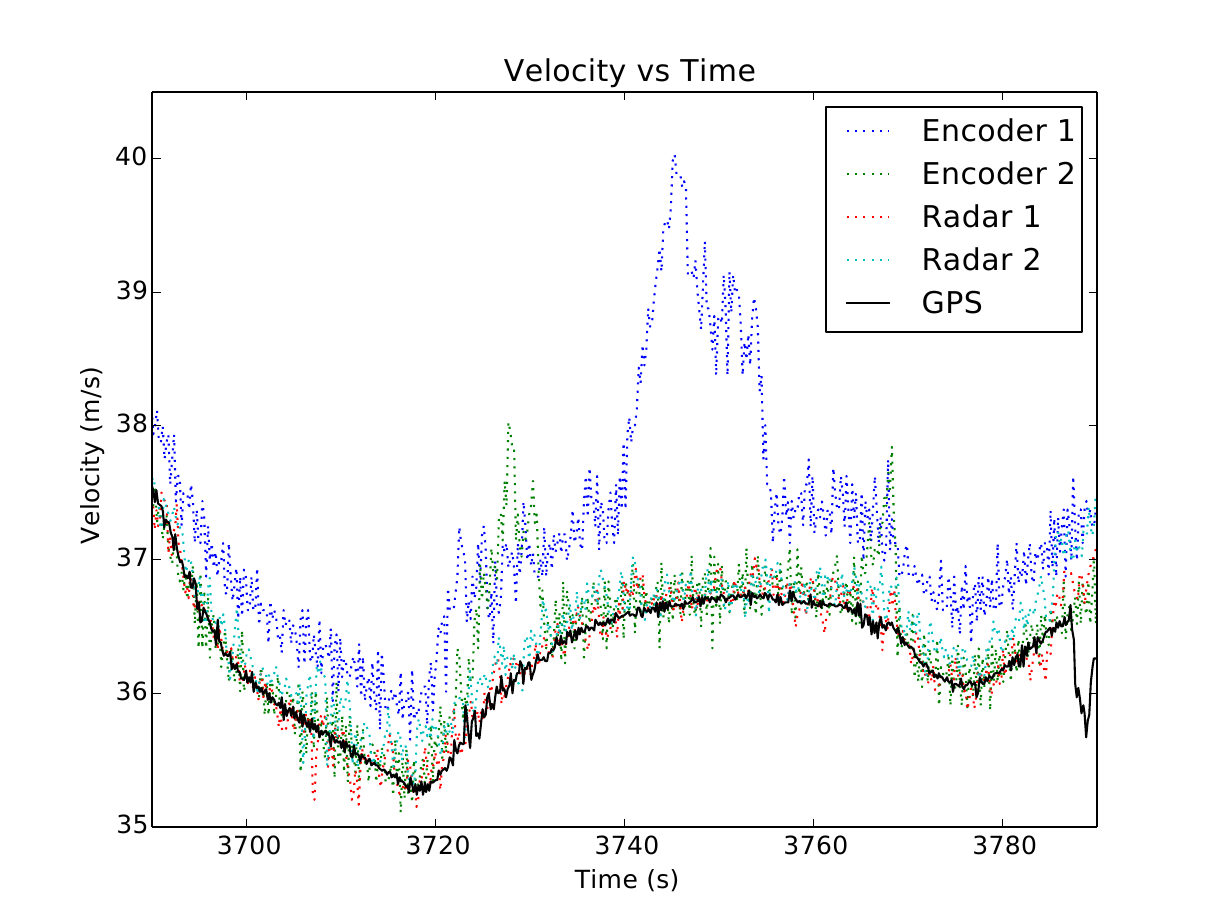}\label{sf:acceleration}
	}
	
	\subfloat[Wheels sliding during braking]{
		\includegraphics[width=\linewidth]{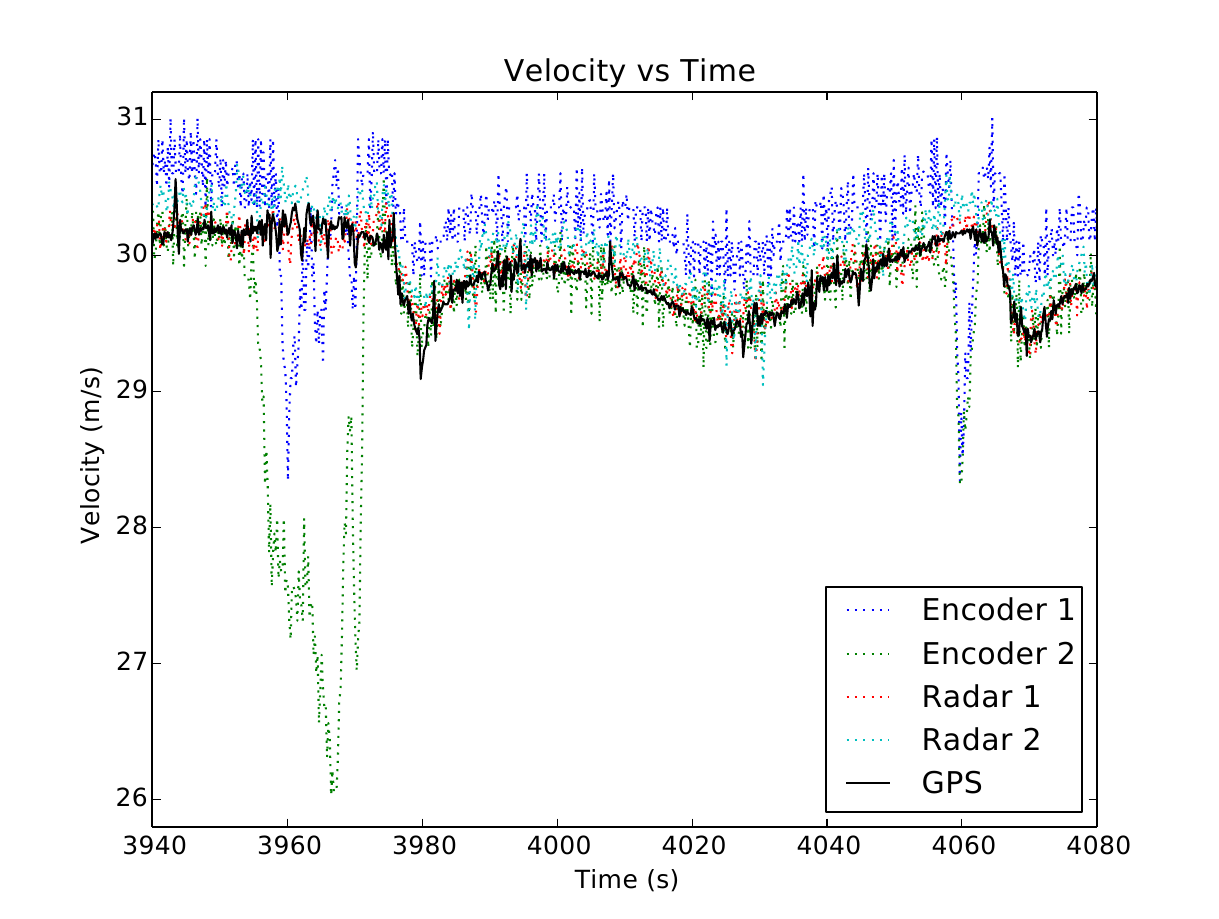}\label{sf:braking}
	}
	
	\subfloat[Frequenct loss of GPS signal]{
		\includegraphics[width=\linewidth]{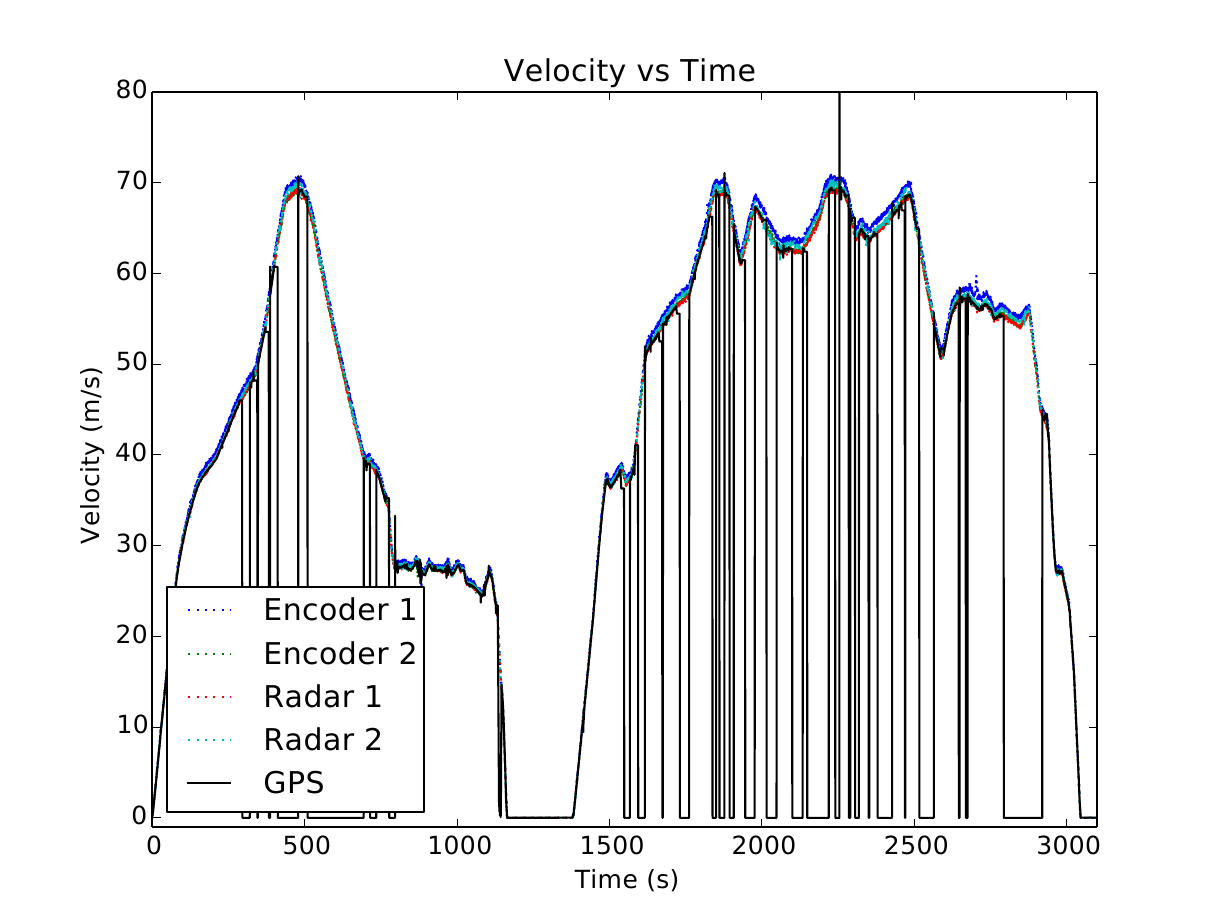}\label{sf:gps}
	}
\caption{Example measurement data from an ICE train in situations with wheel slip and GPS signal loss. 
In all cases wheel encoder 1 is miscalibrated and reads higher than the reference (GPS) velocity during nominal operation. }
\label{f:sensor_data}
\end{figure}

Examples of recorded sensor data complete with wheel slip and miscalibrated wheel diameters are shown in Fig.~\ref{f:sensor_data}. 
In Fig.~\ref{sf:acceleration}, the wheels slipped during acceleration, while in Fig.~\ref{sf:braking} both wheels slid\footnote{We will use the term slip to cover both cases of slipping and sliding.} during braking. 
Furthermore, wheel encoder one is miscalibrated and consistently reports a higher velocity than the other sensors when not slipping. 
Detecting wheel slip in these examples is challenging, particularly as the acceleration of the wheels due to slip is within the range of possible accelerations of the train. 
Wheel slip is also challenging to deal with using sensor outlier detection methods as it is not characterised by isolated anomalies, but rather the measurements slowly diverge from the true state over a period of time. 
Global Positioning System\footnote{We use the term GPS as a generic term which covers all forms of Global Navigation Satellite Systems (GNSS).} (GPS) can give a slip-free speed measurement. 
However, GPS is particularly unreliable in the rail domain due to the frequent presence of tunnels, cuttings, bridges, and overhanging vegetation. 
This is highlighted in Fig.~\ref{sf:gps}, where GPS signal is lost multiple times for up to several minutes at a time. 

This paper develops an odometry system that deals with the above problems in two ways. 
Firstly, miscalibration of the wheel encoders is handled by incorporating the wheel diameters as states within an Extended Kalman Filter (EKF). 
As a side benefit, this information can be used to automatically signal when wheels require replacement due to wear. 
Secondly, a sensor measurement pre-processing step, called Sensor Consensus Analysis (SCA), is proposed for dealing with untrustworthy sensor readings. 
SCA uses the statistical z-test, along with a user-defined probability, to scale the uncertainties of individual measurements during periods when sensors produce readings that are inconsistent with each other. 
Together, these methods will be shown to produce odometry estimates that are robust to wheel slip and miscalibration errors. 

The rest of this paper is structured as follows: Section~\ref{s:background} presents an overview of relevant literature on odometry, along with a brief look at existing methods for dealing with anomalous sensor measurements when performing filtering. 
Section~\ref{s:ekf}  outlines the proposed EKF formulation that incorporates the calibration of the wheel encoders as state variables. 
Following this, the measurement pre-processing step using the proposed SCA algorithm is presented in Section~\ref{s:consensus}. 
Section~\ref{s:results} demonstrates the efficacy of the proposed methods on data collected from ICE trains. 
Concluding remarks and suggestions for future work are provided in Section~\ref{s:conclusion}.

\section{Background Literature} \label{s:background}

Odometry for rail vehicles has received substantial attention in the research community, with a particular focus on methods for dealing with wheel slip. 
A largely heuristic approach based on comparing accelerations and differences in speeds was developed in \cite{Allotta2001} for determining when wheel slip was occurring. 
This approach was compared against fuzzy logic and neural network approaches in \cite{Allotta2002} and \cite{Colla2003}. 
The authors noted that, while the fuzzy logic and neural network approaches provided superior results to the heuristic approach, the difficulty in formally verifying the safety of these approaches hinders their adoption by industry. 
Their heuristic approach was refined in \cite{Ridolfi2013} with the addition of Inertial Measurement Units (IMUs) to aid in the detection of wheel slip. 
Slipping wheels were detected by comparing the acceleration of the wheel with the measured acceleration from the IMU, with the affected measurements subsequently discarded. 
While the above approaches produce good results, they lack any measure quantifying the confidence or uncertainty of the odometry estimate due to their heuristic nature. 

Other work in this domain has focused on developing sensing methods that are not affected by wheel slip. 
Bah et al. 
\cite{Bah2009} analysed a Visual Odometry (VO) system that used only a front-facing video camera to estimate the speed of the train by matching features between successive frames. 
While the speed estimate using only two frames was quite noisy, once averaged over several readings they found that it was accurate to within approximately 3.7\%. 
An eddy-current sensor was developed in \cite{Engelberg2000}. 
This type of sensor was used in \cite{Lauer2015} to provide a slip-free velocity estimate as an input to a map-based localization system. 
Another advantage of eddy-current sensors over wheel encoders and Doppler radars is that they can also be used to detect features in the railway, such as switches, for use in localization algorithms. 
The downside of using additional sensors such as IMUs, cameras, and eddy-current sensors, is that retrofitting them to existing trains is expensive, due not only to the large number of trains in operation, but also the extensive safety certification process required. 
Therefore, this paper investigates methods of improving the odometry algorithm without requiring the addition of new sensors. 

Outside of the rail domain, common approaches for dealing with wheel slip on robots and other vehicles include detecting wheel slip using the motor current \cite{Ojeda2006}, machine learning approaches \cite{Iagnemma2009}, or a detailed dynamic model \cite{Ward2008}, and using alternative sensing methods such as VO \cite{Helmick2004, Nister2006, Maimone2007, Nourani-Vatani2011}. 
Combining VO with motion models that accurately model the dynamic constraints of the vehicle yielded good results in the above literature. 
Using dynamic constraints (by, for example, using a constrained Kalman filter \cite{Palmer2016} to limit the acceleration range of the state estimate) could provide some benefit for railway applications. 
However, as previously noted, the accelerations seen in the data due to wheel slip are generally within the range of expected accelerations of the trains, minimising the benefit of such an approach. 

In practice, sensors are not perfect and can produce anomalous readings for a multitude of reasons. 
Outlier rejection methods are commonly used to detect and remove such measurements. 
A common approach for detecting outlier measurements in Kalman filter based estimation systems is through a chi-squared distribution test \cite{Sukkarieh1999}, also known as the Mahalanobis distance or Normalized Innovation Squared (NIS) \cite{Franke2005,Morales2008}. 
If a measurement fails the chi-squared test against the current estimated state for a user-defined probability threshold, the measurement is discarded. 
As noted in \cite{Worrall2015}, this approach can work well for isolated error spikes, but does not work for long periods where the sensors produce anomalous readings. 



\section{Extended Kalman Filter} \label{s:ekf}

This section develops an EKF for tracking the state of the train. 
The state, $\boldsymbol{x}$, contains the distance driven, the velocity, and acceleration of the train, along with calibration factors for tracking changes in the wheel diameters used for deriving velocity measurements from the wheel encoders:
\begin{equation} \label{eq:state}
\boldsymbol{x} = \begin{bmatrix}
\text{distance} \\
\text{velocity} \\
\text{acceleration} \\
\text{encoder 1 calibration} \\
\text{encoder 2 calibration} \\
\end{bmatrix}.
\end{equation}

The state evolves according to the following linear model:
\begin{equation}
\boldsymbol{x}_{k} = \boldsymbol{F}_{k}\boldsymbol{x}_{k-1} + \boldsymbol{B}_{k}\boldsymbol{u}_{k} + \boldsymbol{w}_{k},
\end{equation}
where $\boldsymbol{F}_{k}$ is the state transition model, $\boldsymbol{B}_{k}$ is the control-input model, $\boldsymbol{u}_{k}$ is the control-input vector, and $\boldsymbol{w}_{k}$ is the process noise. 
For the state vector given in \eqref{eq:state}, the state transition model used was
\begin{equation}
\boldsymbol{F}_{k} = \begin{bmatrix}
1 & \Delta t & \frac{\left(\Delta t\right)^{2}}{2} & 0 & 0 \\
0 & 1 & \Delta t & 0 & 0 \\
0 & 0 & 1 & 0 & 0 \\
0 & 0 & 0 & 1 & 0 \\
0 & 0 & 0 & 0 & 1 
\end{bmatrix}.
\end{equation}
where $\Delta t$ is the time difference to the previous state. 

The trains under consideration do not provide an interface for reading the control inputs, so the control-input model and vector were not required. The measurement model for radar measurements is linear:
\begin{equation}
\boldsymbol{m}_{k} = \boldsymbol{H}_{k}\boldsymbol{x}_{k} + \boldsymbol{v}_{k},
\end{equation}
where $\boldsymbol{m}_{k}$ is the measurement of the true state $\boldsymbol{x}_{k}$, $\boldsymbol{H}_{k}$ is the observation model, and $\boldsymbol{v}_{k}$ is the observation noise. 
The observation model for radar measurements is simply
\begin{equation}
\boldsymbol{H}_{k} = \begin{bmatrix}
0 & 1 & 0 & 0 & 0
\end{bmatrix},
\end{equation}
as the measurement from a Doppler radar is a direct measurement of the velocity. 
For the wheel encoders, the measurement function is a non-linear combination of the states. 
For a measurement from encoder 1,
\begin{align}
\boldsymbol{m}_{k} &= h\left(\boldsymbol{x}_{k}\right) + \boldsymbol{v}_{k} \\
&= \begin{bmatrix}
\frac{\boldsymbol{x}_{k,1}}{\boldsymbol{x}_{k,3}}
\end{bmatrix} + \boldsymbol{v}_{k},
\end{align}
where $\boldsymbol{x}_{k,1}$ is the velocity from the state vector $\boldsymbol{x}_{k}$, and $\boldsymbol{x}_{k,3}$ is the calibration factor for encoder 1. 
This observation model is linearised around the current state to produce the observation matrix
\begin{equation}
\boldsymbol{H}_{k} = \begin{bmatrix}
0 & \frac{1}{\boldsymbol{x}_{k,3}} & 0 & -\frac{\boldsymbol{x}_{k,1}}{\boldsymbol{x}_{k,3}^{2}} & 0
\end{bmatrix}.
\end{equation}

For a measurement from encoder 2, the observation matrix is
\begin{equation}
\boldsymbol{H}_{k} = \begin{bmatrix}
0 & \frac{1}{\boldsymbol{x}_{k,4}} & 0 & 0 & -\frac{\boldsymbol{x}_{k,1}}{\boldsymbol{x}_{k,4}^{2}}
\end{bmatrix}.
\end{equation}

Once these have been calculated, the filter can be run to determine the state estimate, $\hat{\boldsymbol{x}}_{k|k}$, and covariance, $\boldsymbol{P}_{k|k}$. 
Since the state transition model is linear, the standard Kalman Filter equations for the time update step can be used:
\begin{equation}
\hat{\boldsymbol{x}}_{k|k-1} = \boldsymbol{F}_{k}\hat{\boldsymbol{x}}_{k-1|k-1} + \boldsymbol{B}_{k}\boldsymbol{u}_{k},
\end{equation}
\begin{equation}
\boldsymbol{P}_{k|k-1} = \boldsymbol{F}_{k}\boldsymbol{P}_{k|k-1}\boldsymbol{F}_{k}^{T} + \boldsymbol{Q}_{k},
\end{equation}
where $\boldsymbol{Q}_{k}$ is the covariance matrix of the process noise. 
As the observation model can be non-linear, the measurement update equations are
\begin{equation}
\boldsymbol{K}_{k} = \boldsymbol{P}_{k|k-1}\boldsymbol{H}_{k}^{T}\left(\boldsymbol{H}_{k}\boldsymbol{P}_{k|k-1}\boldsymbol{H}_{k}^{T} + \boldsymbol{R}_{k}\right)^{-1},
\end{equation}
\begin{equation}
\hat{\boldsymbol{x}}_{k|k} = \hat{\boldsymbol{x}}_{k|k-1} + \boldsymbol{K}_{k} \left(\boldsymbol{m}_{k} - h\left(\hat{\boldsymbol{x}}_{k|k-1}\right)\right),
\end{equation}
\begin{equation}
\boldsymbol{P}_{k|k} = \left(\boldsymbol{I} - \boldsymbol{K}_{k} \boldsymbol{H}_{k} \right) \boldsymbol{P}_{k|k-1},
\end{equation}
where $\boldsymbol{R}_{k}$ is the covariance matrix of the measurement noise. 

\section{Sensor Consensus Analysis} \label{s:consensus}

SCA is a pre-processing step that exploits the fact that, in the scenario under consideration in this paper, multiple sensors are used to measure the same physical quantity. 
It scales the uncertainty of measurements from individual sensors that are inconsistent with the measurements made at the same time from the other sensors before they are incorporated into the filter, using a statistical z-test to determine whether the uncertainty of a measurement should be scaled or not. 
The idea behind this approach is that the sensor measurements should have significantly overlapping probability distributions when they are measuring the same underlying distribution. 
When this is true, they should pass a z-test with a high probability that the means of the distributions that they are measuring are equal. 
A user specified probability is used to determine the z-test value required as
\begin{equation}
z_{desired} = \text{norminv}\left(1-\frac{p}{2}\right),
\end{equation}
where norminv returns the z-value required for the cumulative distribution function of a standard normal distribution to equal the probability $\frac{p}{2}$, and $z_{desired} \ge 0$. 
Note that the user specified probability, $p$, is for a two-tailed test, which is why it is halved when calculating the z-value.
For any pair of sensor measurements, the z-value corresponding to the probability that the two measurements are of the same population mean is calculated as
\begin{equation}
z_{test} = \frac{|\mu_1 - \mu_2|}{\sqrt{(\sigma_1^2 + \sigma_2^2)}}, 
\end{equation}
where $\mu_i$ and $\sigma_i^2$ are the mean and variance of the measurement from sensor $i$ respectively. 
If $z_{test} > z_{desired}$, then the measurements are not of the same population mean at a significance level $p$.
When a measurement fails this test (i.e., statistically it is measuring a different population or quantity to the measurement being tested against), then the proposed approach is to iteratively scale the uncertainty of the measurements that are in consensus with the least number of the other measurements until all measurements are in consensus with each other. 

The SCA algorithm is outlined in Algorithm~\ref{a:consensus}. 
It takes as input a list of the last received measurement from each sensor, and the user specified probability threshold. 
In each iteration, it calculates the minimum scaling factor required for at least one of the measurements to be in consensus with a measurement it was not previously in consensus with, and updates the scaling factors of the measurements with the minimum consensus. 
The uncertainty of each measurement is then scaled in the \texttt{ScaleMeasurements} function by multiplying its variance by the scaling factor for that measurement, before starting the next iteration. 

\begin{algorithm}[tbp]
\SetKwInOut{Input}{input}
\SetKwInOut{Result}{result}
\SetKwFunction{CalculateMinScale}{CalculateMinScale}
\SetKwFunction{ScaleMeasurements}{ScaleMeasurements}
\SetKwFunction{length}{length}

\Input{List of measurements, $\boldsymbol{M}$, consensus probability, $p$}
\Result{List of scaling factors for each measurement, $\boldsymbol{S}$}
\BlankLine

$\boldsymbol{S} = [1$ \textbf{for} $i \in \{0,\dots,$\length{$\boldsymbol{M}$}$-1]$\;
$\boldsymbol{N} = \boldsymbol{M}$\;
\While{at least one pair of measurements in $\boldsymbol{N}$ do not pass the consensus test with probability, $p$} {
	$\boldsymbol{L}$ = list of indexes of measurements in $\boldsymbol{N}$ with the minimum consensus\;
	$c$ = \CalculateMinScale{$\boldsymbol{L},\boldsymbol{N},p$}\;
	\For{$i \in \boldsymbol{L}$}{
		$\boldsymbol{S}[i] = \boldsymbol{S}[i] \times c$\;
	}
	$\boldsymbol{N} = $ \ScaleMeasurements{$\boldsymbol{M},\boldsymbol{S}$}\;
}
\caption{Sensor consensus analysis}
\label{a:consensus}
\end{algorithm}

The \texttt{CalculateMinScale} function, shown in Algorithm~\ref{a:calc_min_scale}, calculates the minimum scaling factor required for at least one of the measurements listed in $\boldsymbol{L}$ to pass the consensus test with a measurement that it was not previously in consensus with. 
If both measurements under consideration are in the list $\boldsymbol{L}$ (i.e., they are in consensus with the same number of measurements), then both measurements are scaled. 
Consider two measurements $i$ and $j$ with means $\mu_{i}$ and $\mu_{j}$ and variances $\sigma_{i}^{2}$ and $\sigma_{j}^{2}$, and a desired z-value of $z$. 
The scaling factor, $s$, when scaling both measurements is calculated by
\begin{equation}
z = \frac{|\mu_{i} - \mu_{j}|}{\sqrt{s\left(\sigma_{i}^{2} + \sigma_{j}^{2}\right)}}. 
\end{equation}
Rearranging for $s$ gives
\begin{equation}
s = \frac{\left(\mu_{i} - \mu_{j}\right)^{2}}{z^{2}\left(\sigma_{i}^{2} + \sigma_{j}^{2}\right)}.
\end{equation}

If measurement $j$ is not in $\boldsymbol{L}$, then only measurement $i$ is scaled when calculating the scaling factor:
\begin{equation}
z = \frac{|\mu_{i} - \mu_{j}|}{\sqrt{s\sigma_{i}^{2} + \sigma_{j}^{2}}}.
\end{equation}
Rearranging for $s$ gives
\begin{equation}
s = \frac{\left(\frac{\mu_{i} - \mu_{j}}{z}\right)^{2} - \sigma_{j}^{2}}{\sigma_{i}^{2}}.
\end{equation}

\begin{algorithm}[tbp]
\SetKwInOut{Input}{input}
\SetKwInOut{Result}{result}
\SetKwFunction{length}{length}
\SetKwFunction{CalcScaleBoth}{CalcScaleBoth}
\SetKwFunction{CalcScaleOne}{CalcScaleOne}

\Input{List of measurements to scale, $\boldsymbol{L}$, list of measurements, $\boldsymbol{N}$, desired probability, $p$}
\Result{The minimum scaling factor, $c$}
\BlankLine

$c = \infty$\;
$z = \texttt{norminv}\left(\frac{p}{2}\right)$\;
\For{$i \in \boldsymbol{L}$}{
	\For{$j \in \{0,\dots,$\length{$\boldsymbol{N}$}$-1\}$}{
		\If{$\boldsymbol{N}[i]$ is not in consensus with $\boldsymbol{N}[j]$}{
			\eIf{$j \in \boldsymbol{L}$}{
				$s = $\CalcScaleBoth{$\boldsymbol{N}[i],\boldsymbol{N}[j],z$}\;
			}{
				$s = $\CalcScaleOne{$\boldsymbol{N}[i],\boldsymbol{N}[j],z$}\;
			}
			\If{$s < c$}{
			$c = s$\;
			}
		}
	}
}
\caption{CalculateMinScale}
\label{a:calc_min_scale}
\end{algorithm}

\begin{figure}[htbp]
	\centering
	\subfloat[Input measurements]{
		\includegraphics[width=0.76\linewidth]{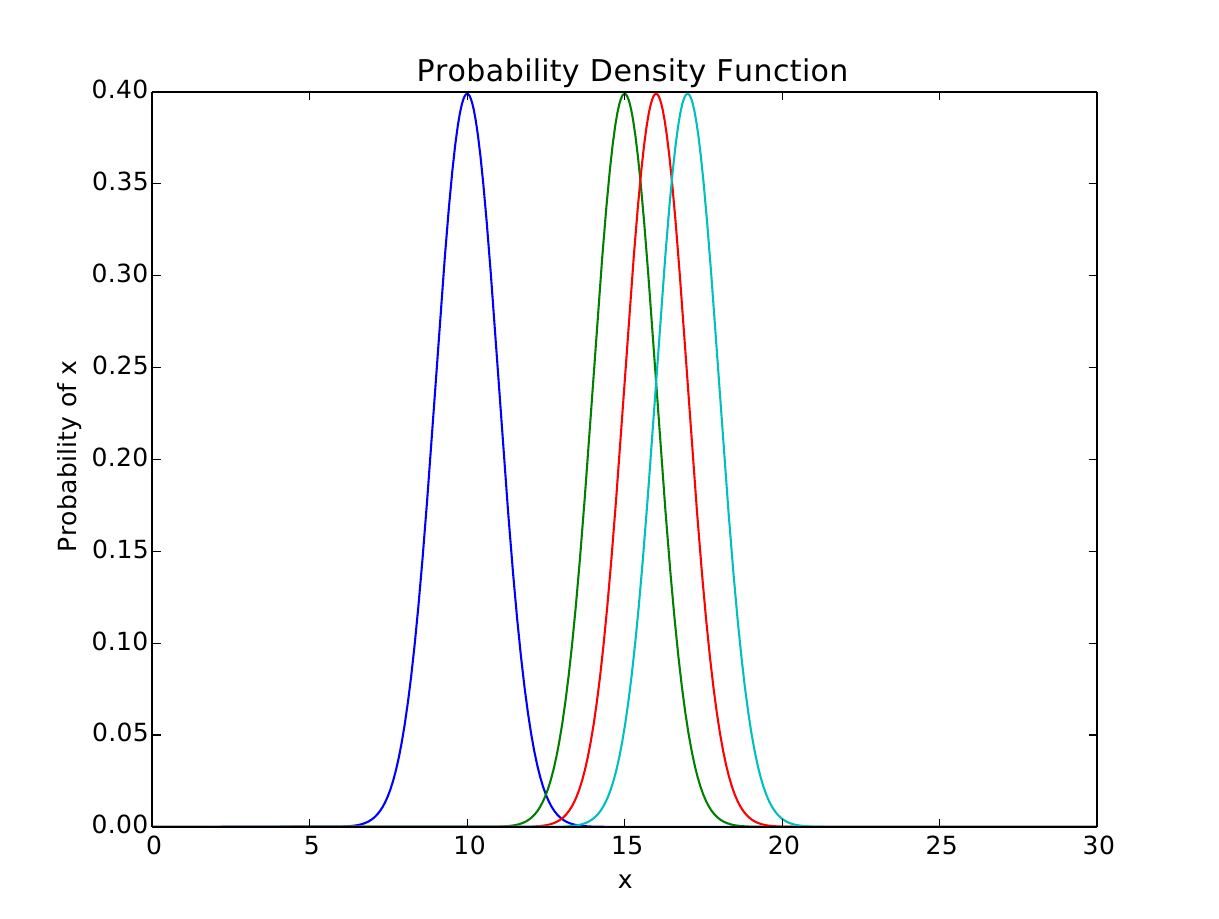}
	}

	\subfloat[Scaled measurements]{
		\includegraphics[width=0.76\linewidth]{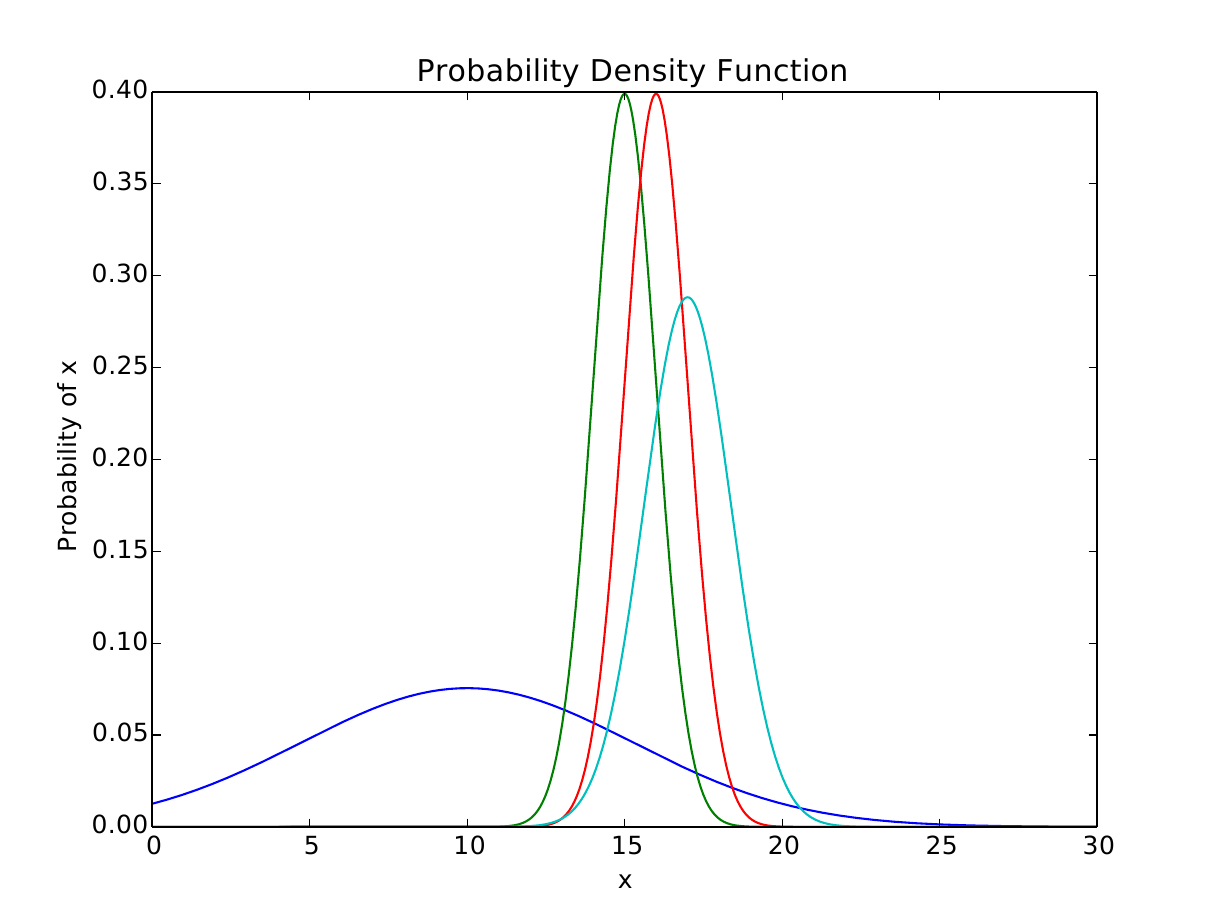}
	}
\caption{SCA example with 1 outlier measurement. }
\label{f:consensus_example_1}
\end{figure}

\begin{figure}[htbp]
	\centering
	\subfloat[Input measurements]{
		\includegraphics[width=0.76\linewidth]{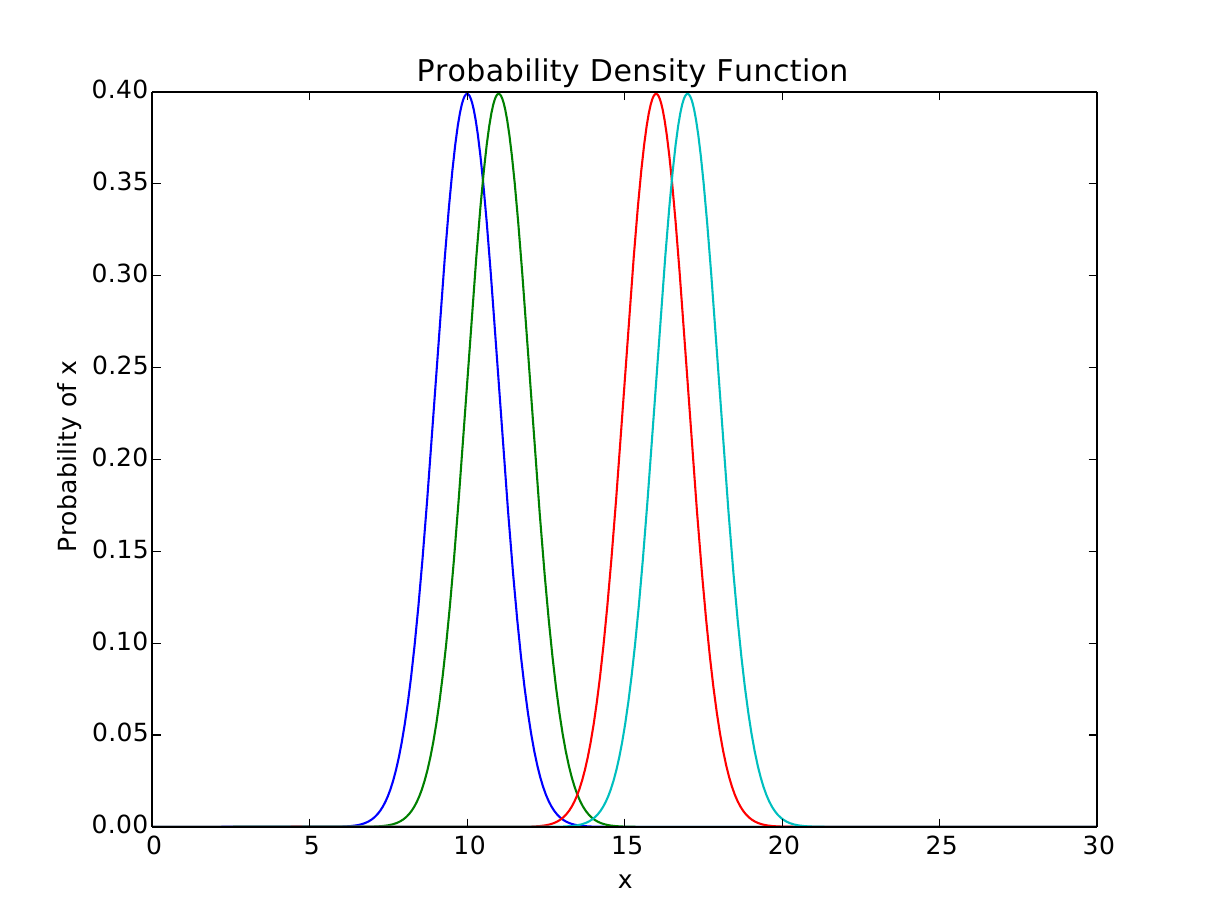}
	}
	
	\subfloat[Scaled measurements]{
		\includegraphics[width=0.76\linewidth]{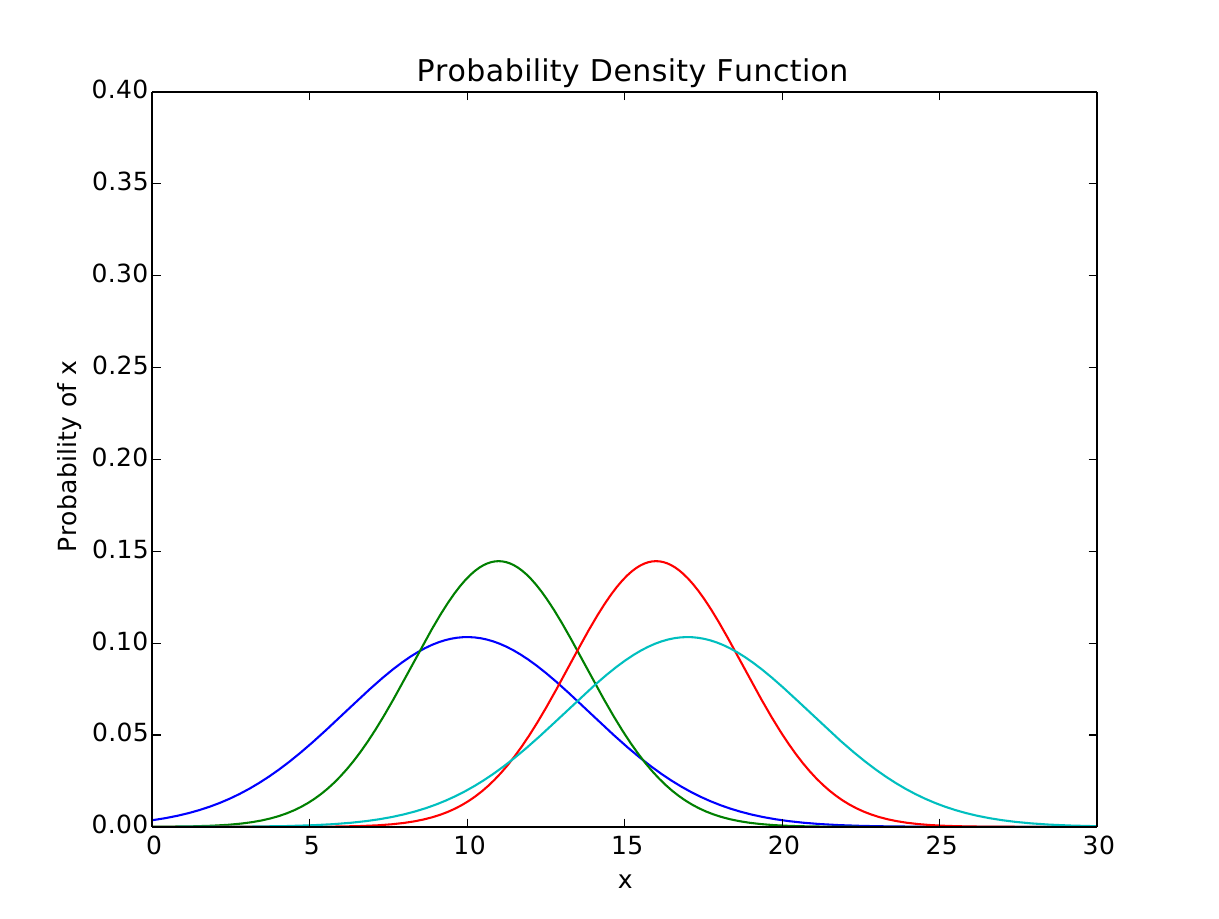}
	}
\caption{SCA example with two groups of measurements. }
\label{f:consensus_example_2}
\end{figure}

Fig.~\ref{f:consensus_example_1} and \ref{f:consensus_example_2} show two examples of SCA applied to a set of four measurements from different sensors, using a probability threshold of 0.2. 
In the first example, three of the measurements are grouped together, resulting in the one inconsistent measurement being scaled significantly. 
Since measurements $b$, $c$ and $d$ were already in consensus with one another, they do not receive as much scaling. 
In the second example, there are two groups of paired measurements, resulting in all measurements being scaled substantially. 

\section{Results} \label{s:results}

\begin{figure*}[htbp]
	\centering
	\subfloat[Results without calibration]{
		\includegraphics[width=0.48\linewidth]{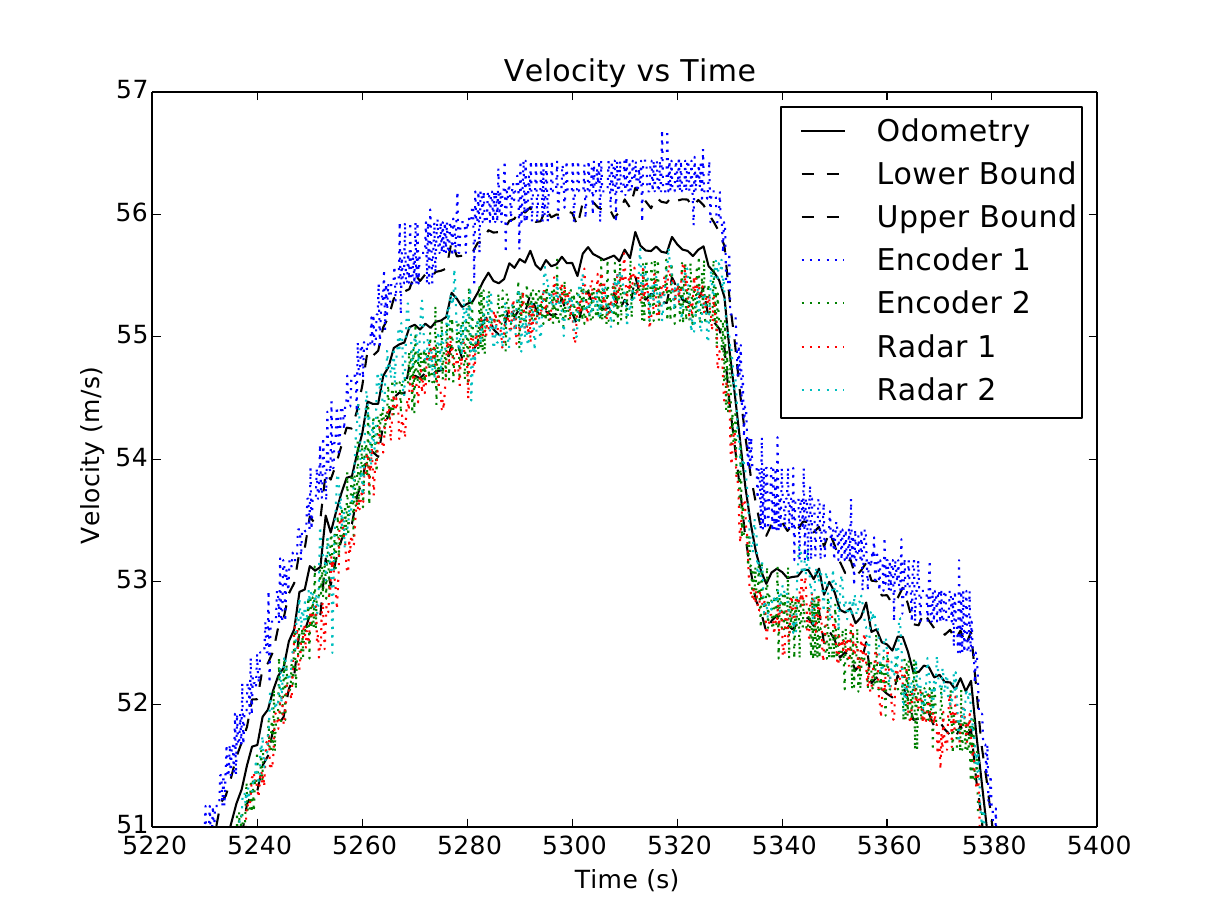}
	}
	\subfloat[Results with calibration]{
		\includegraphics[width=0.48\linewidth]{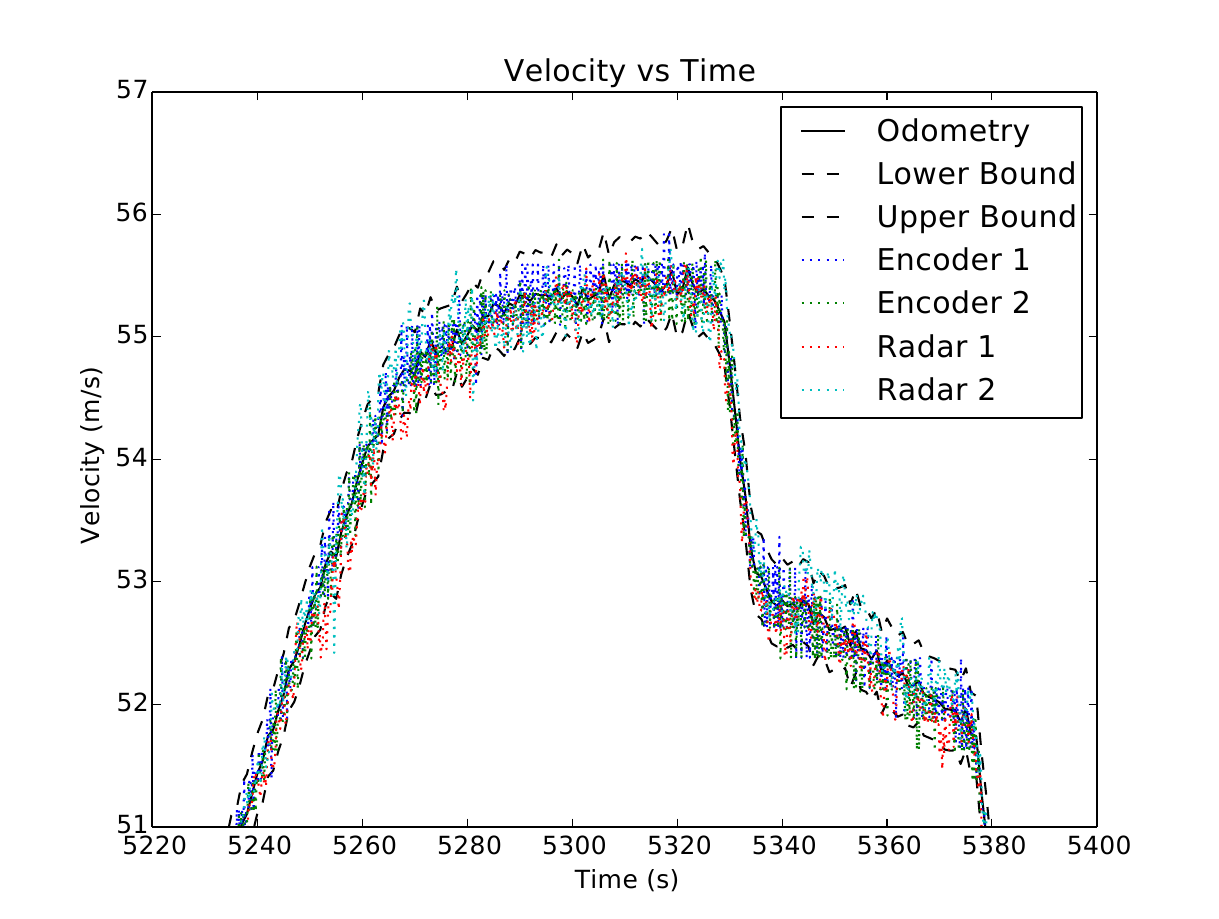}
	}
\caption{Successful calibration of the wheel diameter. The upper and lower bounds of the odometry estimate are $\mu \pm \sigma$.}
\label{f:calibration}
\end{figure*}

The proposed methods were tested on data collected from an ICE train operating in Germany, which  was equipped with two Doppler radars and two wheel encoders, each reporting measurements at 5Hz. 
The wheel encoders report the cumulative ticks during the measurement period. 
The on-line wheel diameter calibration was first evaluated. 
As can be seen in Fig.~\ref{f:calibration}, the EKF successfully calibrates the wheel encoders, aligning them with the Doppler radars. 
It is important that the process noise of the calibration values are small to ensure that the calibration values are unaffected by wheel slip. 
When the process noise of the calibration values is too large, differences between the radar and wheel encoder measurements are compensated for instantly, even during wheel slip, which is undesirable as it effectively removes the wheel encoders as an information source for the EKF.

%
%

\begin{figure*}[htbp]
	\centering
	\subfloat[$p=0$ (no outlier rejection)]{
		\includegraphics[width=0.48\linewidth]{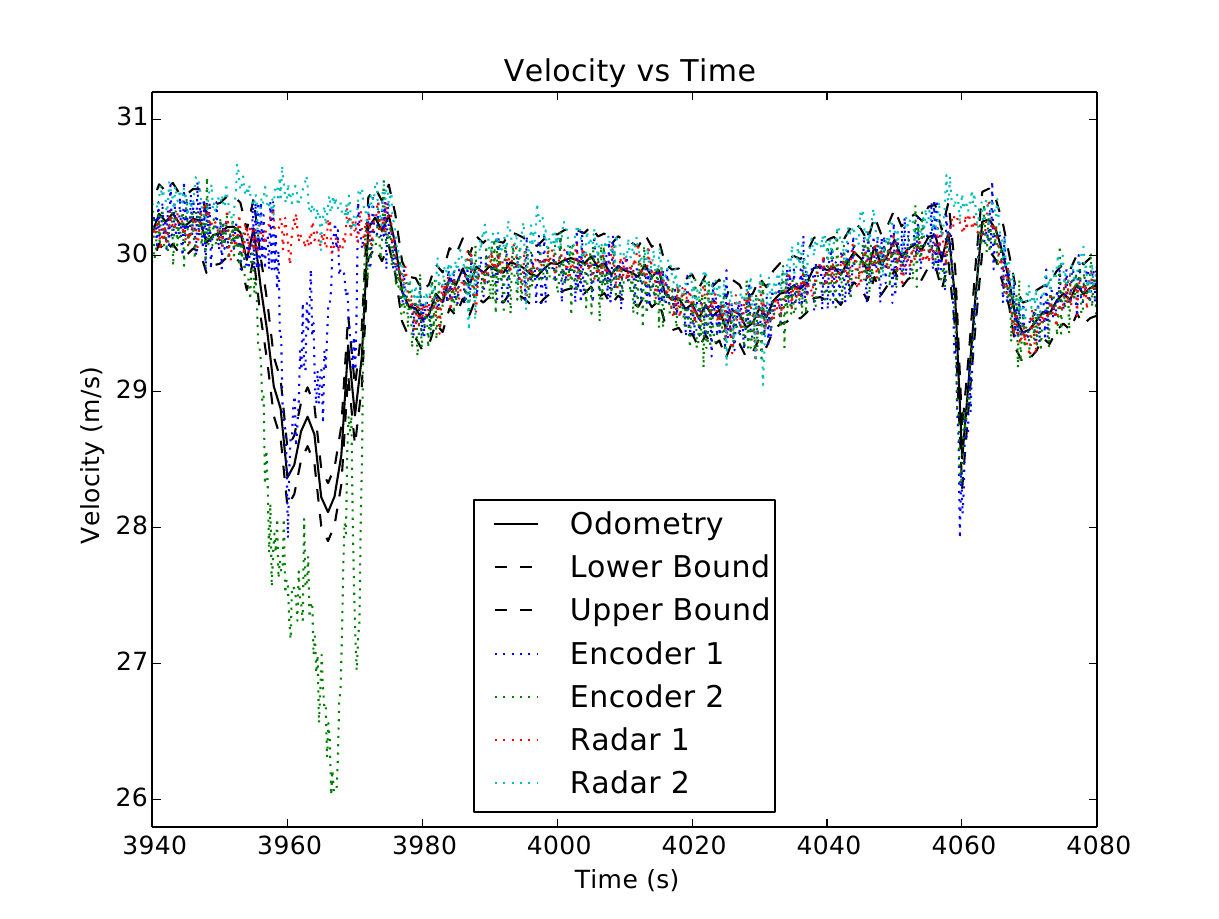}\label{sf:results_no}
	}
	\subfloat[Mahalanobis distance threshold of 3]{
		\includegraphics[width=0.48\linewidth]{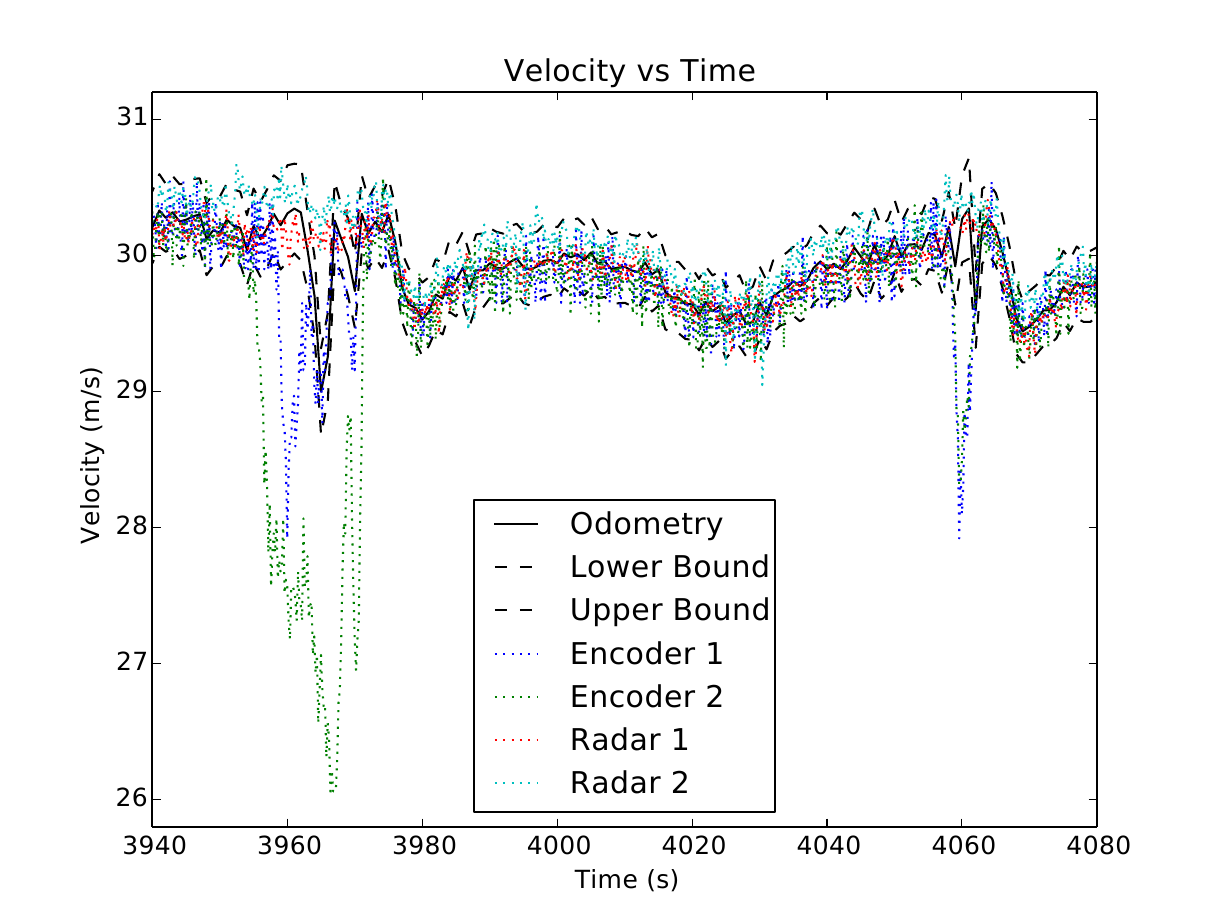}\label{sf:results_mah}
	}
	
	\subfloat[$p=0.5$]{
		\includegraphics[width=0.48\linewidth]{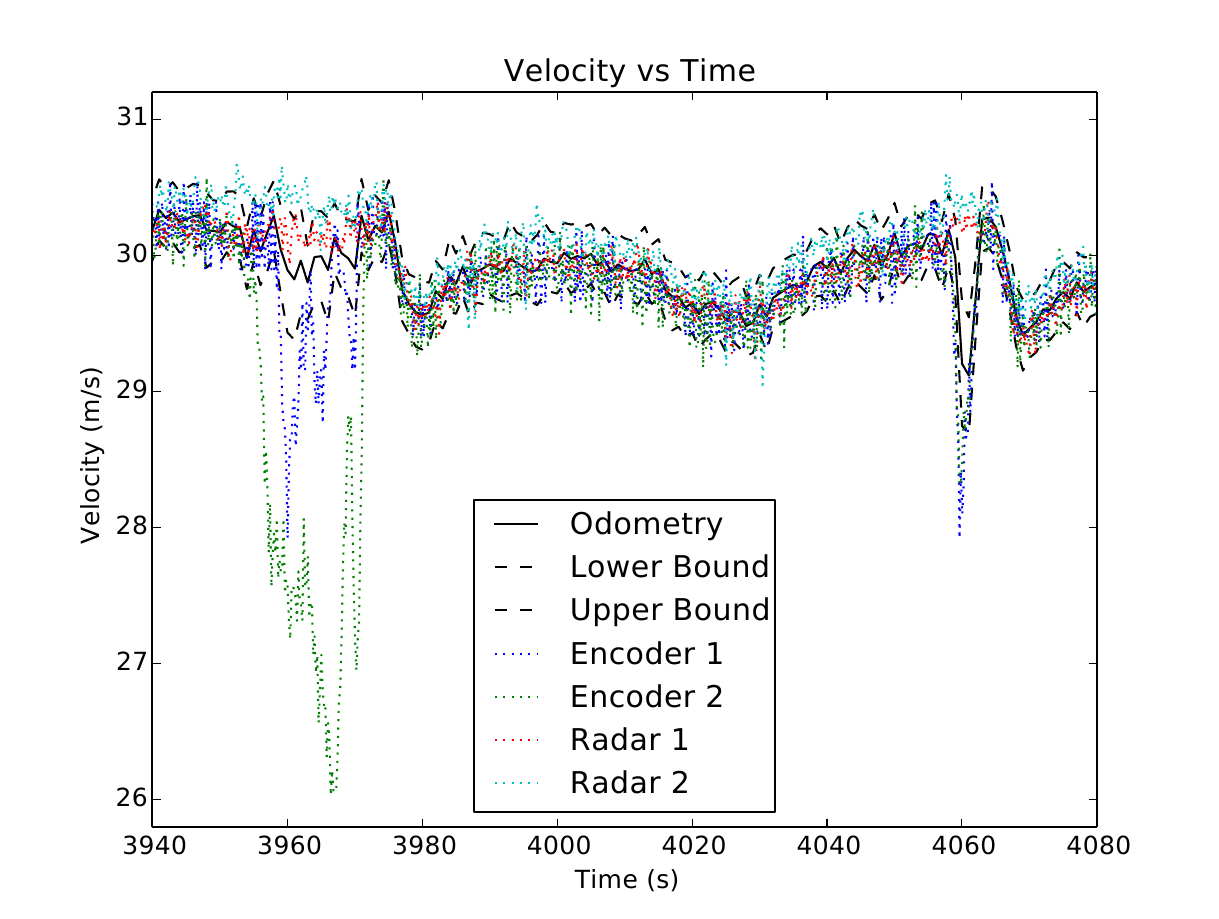}\label{sf:results05}
	}
	\subfloat[$p=0.9$]{
		\includegraphics[width=0.48\linewidth]{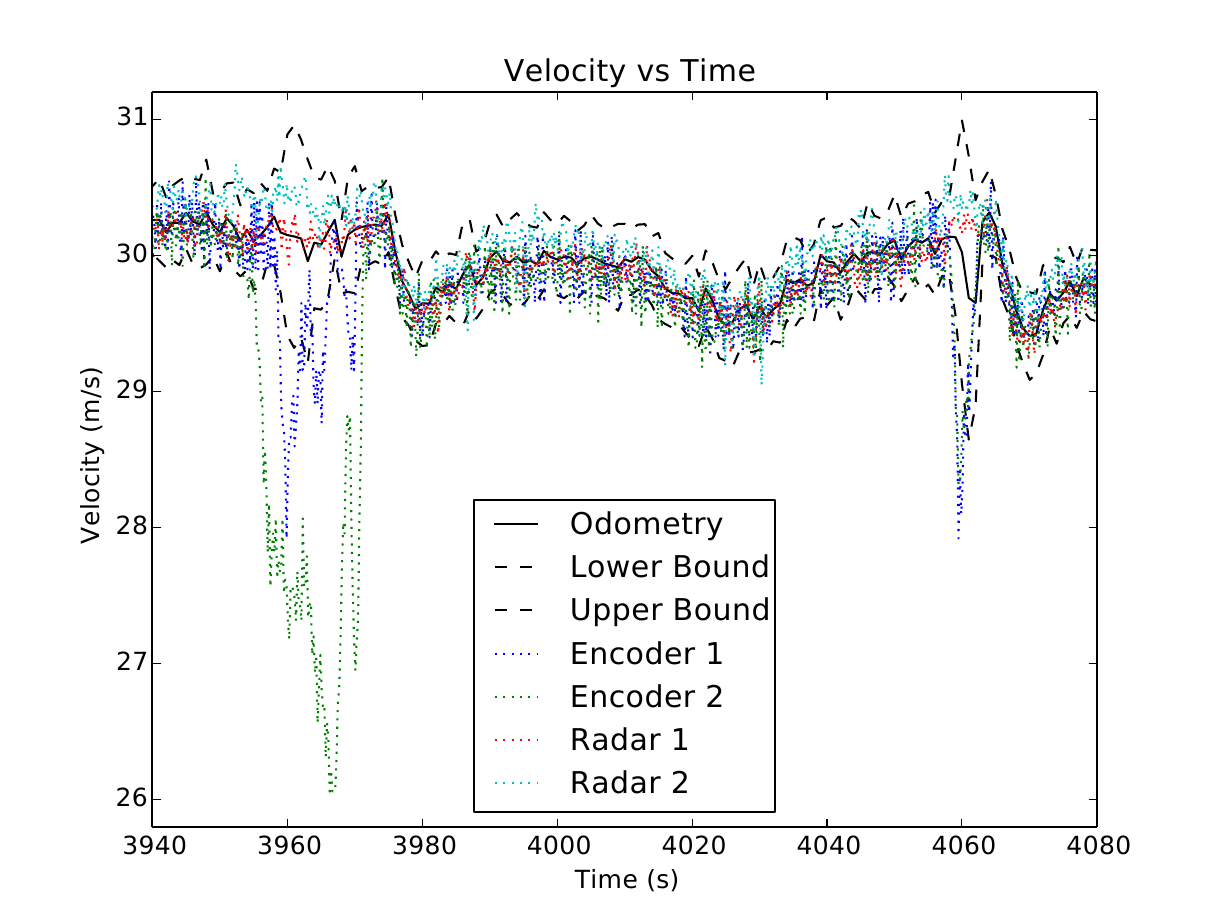}\label{sf:results09}
	}

\caption{Results on an example with two wheel slip events. 
The reference velocity, as given by GPS, can be seen in Fig.~\ref{sf:braking}. 
The upper and lower bounds of the odometry estimate are $\mu \pm \sigma$.}
\label{f:results}
\end{figure*}

Fig.~\ref{f:results} shows the behaviour of the filter with the various measurement pre-processing approaches during an example with two wheel slip events. 
In the first slip event, both wheels experience different amounts of slip, while in the second they experience almost exactly the same amount of slip. 
Without any pre-processing (Fig.~\ref{sf:results_no}), the odometry is significantly affected by the wheel slip. 
Using a low Mahalanobis threshold to remove outliers (Fig.~\ref{sf:results_mah}) yields a performance increase over no pre-processing, but the resultant odometry is still affected by the wheel slip. 
With a probability threshold of 0.5 (Fig.~\ref{sf:results05}), SCA almost entirely removes the effects of the first wheel slip, but is ineffective for the second wheel slip. 
Increasing the threshold to 0.9 (Fig.~\ref{sf:results09}) yields good performance in both cases. 
The second wheel slip event in this case highlights a desirable property of SCA---in this case, there are two distinct groups of measurements, and, without additional information, it is impossible to know which group of measurements is correctly measuring the true velocity of the train. 
As a result, SCA increases the uncertainty of all measurements during the slip event, resulting in a corresponding increase in the uncertainty of the odometry, without significantly affecting the uncertainty outside of the wheel slip event. 
The robustness of the proposed approach was evaluated over 45 minutes of travel using the GPS velocity as the reference velocity, with periods of poor satellite coverage excluded. 
The percentage of time that the GPS velocity was within the 1--$\sigma$ and 3--$\sigma$ bounds of the odometry for various combinations of calibration and measurement pre-processing is shown in Table \ref{tab:results}. 
While the increase in performance of SCA over the Mahalanobis distance at the 3--$\sigma$ level is only 0.15\%, such increases are critical for satisfying the stringent Safety Integrity Level (SIL) requirements of high-speed trains. 

\begin{table}[tbp]
	\caption{Percentage of time that the GPS velocity was within the $n$--$\sigma$ bounds of the odometry}
\begin{center}
\begin{tabular}{c|c||c|c}
	\hline
	\textbf{Calibration} & \textbf{Pre-processing} & \textbf{\% within 1--$\sigma$} & \textbf{\% within 3--$\sigma$} \\
	\hline
	No & None & 46.76\% & 98.03\% \\
	Yes & None & 92.14\% & 99.10\% \\
	Yes & Mahalanobis 3 & 92.30\% & 99.83\% \\
	Yes & SCA 0.5 & 93.81\% & 99.57\% \\
	Yes & SCA 0.9 & \textbf{98.17\%} & \textbf{99.98\%} \\
	\hline
\end{tabular}
\end{center}
\label{tab:results}
\end{table}

%
%

\begin{figure}[]
	\centering
	\subfloat[Mahalanobis distance threshold of 3]{
		\includegraphics[width=0.96\linewidth]{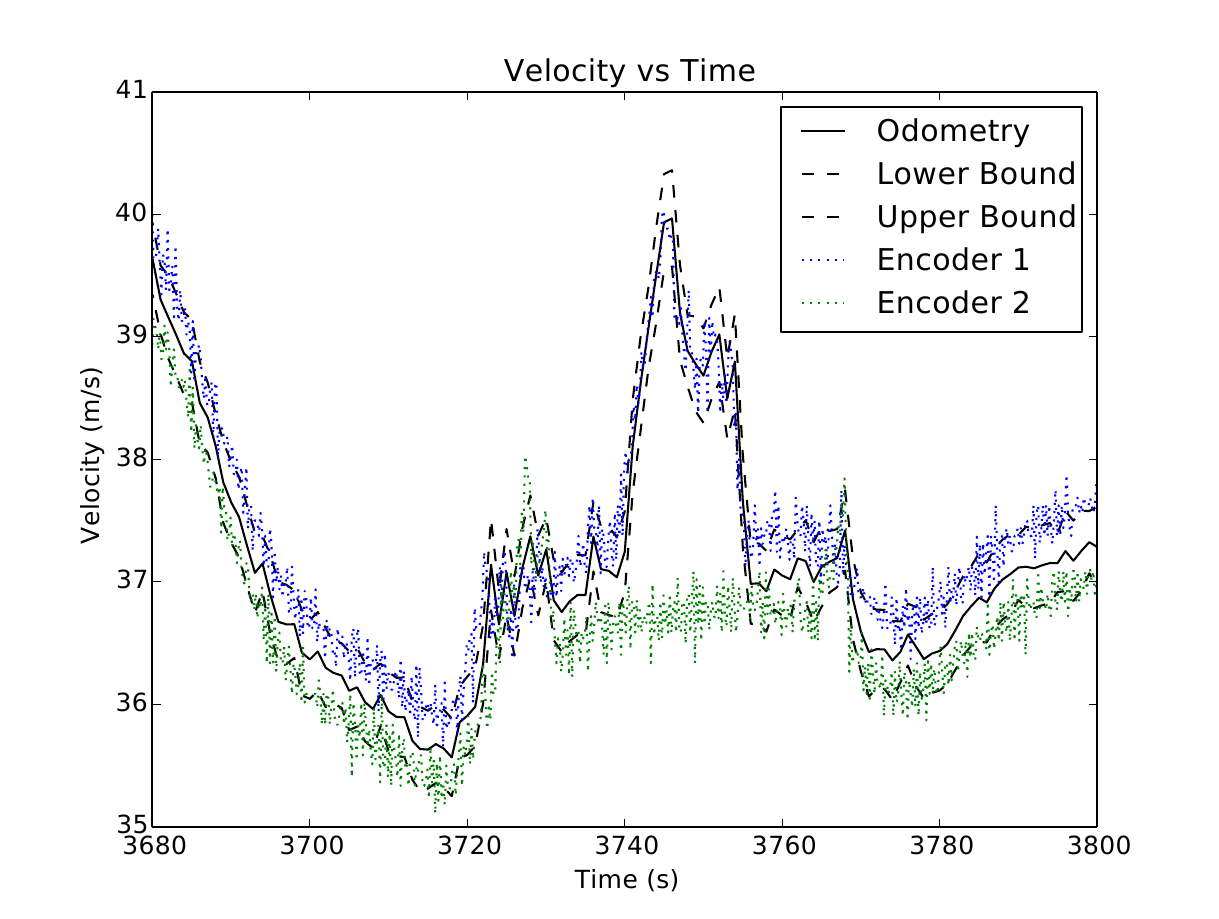}
	}
	
	\subfloat[SCA with $p = 0.8$]{
		\includegraphics[width=0.96\linewidth]{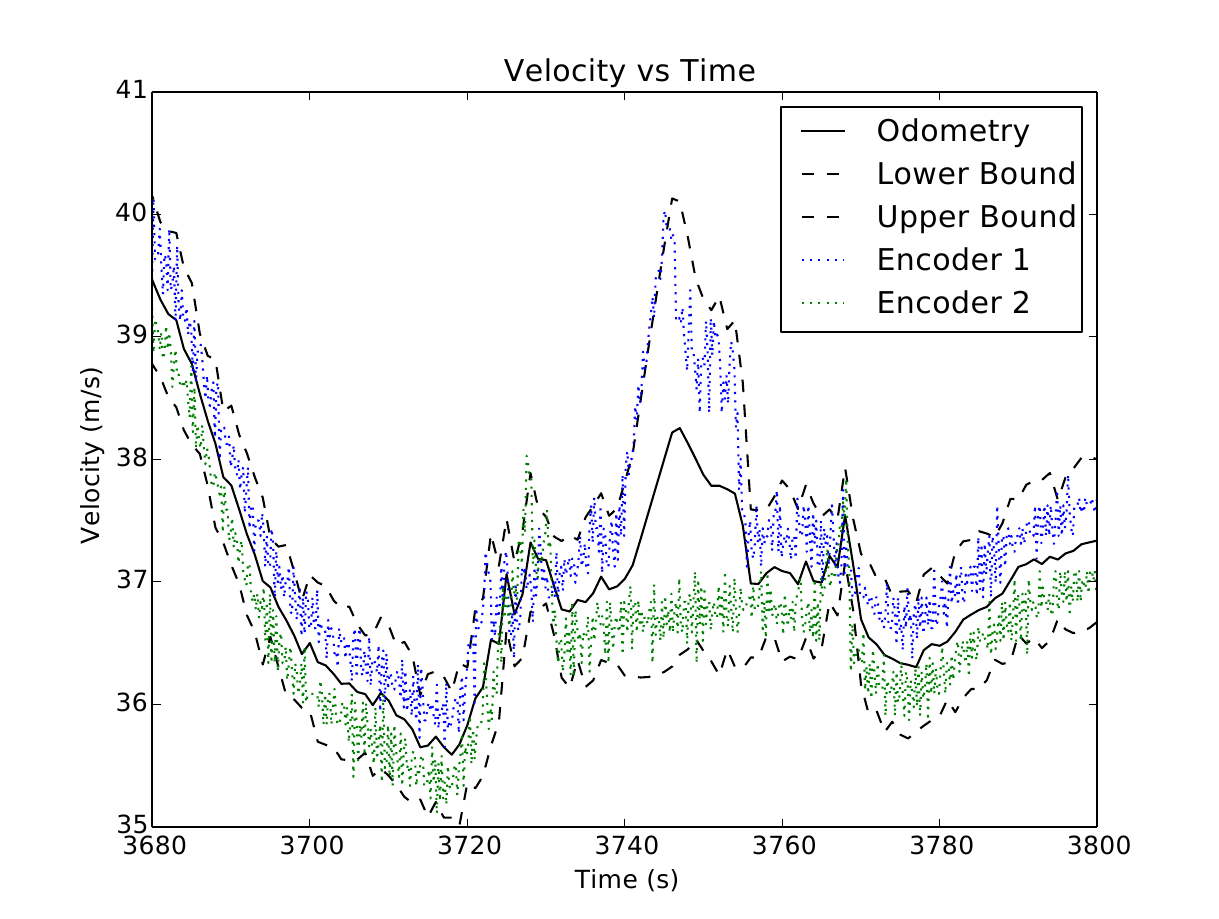}
	}
\caption{Results using only wheel encoders. 
The upper and lower bounds of the odometry estimate are $\mu \pm \sigma$.}
\label{f:only_opg}
\end{figure}

One of the situations that the odometry system must commonly handle is the loss of the Doppler radars due to snow and ice, and hence relying only on the wheel encoders to calculate the velocity of the train. 
Results from such a scenario when wheel slip occurs are presented in Fig.~\ref{f:only_opg}. 
As can be seen, using the Mahalanobis distance to discard outliers in this case results in the EKF closely following one of the diverging sensors at the expense of the other. 
In comparison, the EKF using SCA has increased uncertainty during times when the two sensors diverge, reflecting the uncertainty over which sensor can be trusted. 

\section{Conclusion} \label{s:conclusion}

This paper investigated the problem of robust odometry in the presence of wheel slip and calibration errors, particularly for rail vehicles. 
Calibration errors were successfully dealt with by incorporating the calibration as a state in an EKF. 
Wheel slip and other measurement anomalies were handled through a measurement pre-processing stage called Sensor Consensus Analysis. 
This pre-processing stage inflated the uncertainty of measurements deemed to be inconsistent with the measurements from other sensors. 
The proposed approach was tested on data from German ICE trains, with the benefit of the on-line calibration and SCA clearly demonstrated. 
Ideas for future work include investigating whether SCA can be used with the estimated state as an input (similar to traditional outlier rejection methods), tracking accelerations of the individual velocity sensors and using these as additional inputs to SCA for determining consistency, and incorporating a dynamic model of the vehicle, which includes acceleration limits, into the EKF.



\bibliographystyle{IEEEtran}
\bibliography{IEEEabrv,ref}

\end{document}